\PassOptionsToPackage{unicode}{hyperref}
\PassOptionsToPackage{hyphens}{url}
\PassOptionsToPackage{dvipsnames,svgnames,x11names}{xcolor}
\documentclass[
  american,
  11pt,
  letterpaper,
  11pt]{article}
\usepackage{xcolor}
\usepackage[margin=1in]{geometry}
\usepackage{amsmath,amssymb}
\usepackage{iftex}
\ifPDFTeX
  \usepackage[T1]{fontenc}
  \usepackage[utf8]{inputenc}
  \usepackage{textcomp} 
\else 
  \usepackage{unicode-math} 
  \defaultfontfeatures{Scale=MatchLowercase}
  \defaultfontfeatures[\rmfamily]{Ligatures=TeX,Scale=1}
\fi
\usepackage{lmodern}
\ifPDFTeX\else
\fi
\IfFileExists{upquote.sty}{\usepackage{upquote}}{}
\IfFileExists{microtype.sty}{
  \usepackage[]{microtype}
  \UseMicrotypeSet[protrusion]{basicmath} 
}{}
\usepackage{setspace}
\makeatletter
\@ifundefined{KOMAClassName}{
  \IfFileExists{parskip.sty}{%
    \usepackage{parskip}
  }{
    \setlength{\parindent}{0pt}
    \setlength{\parskip}{6pt plus 2pt minus 1pt}}
}{
  \KOMAoptions{parskip=half}}
\makeatother
\usepackage{longtable,booktabs,array}
\usepackage{calc} 
\usepackage{etoolbox}
\makeatletter
\patchcmd\longtable{\par}{\if@noskipsec\mbox{}\fi\par}{}{}
\makeatother
\IfFileExists{footnotehyper.sty}{\usepackage{footnotehyper}}{\usepackage{footnote}}
\makesavenoteenv{longtable}
\usepackage{graphicx}
\makeatletter
\newsavebox\pandoc@box
\newcommand*\pandocbounded[1]{
  \sbox\pandoc@box{#1}%
  \Gscale@div\@tempa{\textheight}{\dimexpr\ht\pandoc@box+\dp\pandoc@box\relax}%
  \Gscale@div\@tempb{\linewidth}{\wd\pandoc@box}%
  \ifdim\@tempb\p@<\@tempa\p@\let\@tempa\@tempb\fi
  \ifdim\@tempa\p@<\p@\scalebox{\@tempa}{\usebox\pandoc@box}%
  \else\usebox{\pandoc@box}%
  \fi%
}
\def\fps@figure{htbp}
\makeatother
\NewDocumentCommand\citeproctext{}{}

\makeatletter
 \let\@cite@ofmt\@firstofone
 \def\@biblabel#1{}
 \def\@cite#1#2{{#1\if@tempswa , #2\fi}}
\makeatother
\newlength{\cslhangindent}
\newlength{\csllabelwidth}
\newenvironment{CSLReferences}[2] 
 {\begin{list}{}{%
  \setlength{\itemindent}{0pt}
  \setlength{\leftmargin}{0pt}
  \setlength{\parsep}{0pt}
  \ifodd #1
   \setlength{\leftmargin}{\cslhangindent}
   \setlength{\itemindent}{-1\cslhangindent}
  \fi
  \setlength{\itemsep}{#2\baselineskip}}}
 {\end{list}}
\usepackage{calc}

\newcommand{\CSLLeftMargin}[1]{\parbox[t]{\csllabelwidth}{\strut#1\strut}}
\newcommand{\CSLRightInline}[1]{\parbox[t]{\linewidth - \csllabelwidth}{\strut#1\strut}}

\ifLuaTeX
\usepackage[bidi=basic,shorthands=off]{babel}
\else
\usepackage[bidi=default,shorthands=off]{babel}
\fi
\ifLuaTeX
  \usepackage{selnolig} 
\fi
\providecommand{\tightlist}{%
  \setlength{\itemsep}{0pt}\setlength{\parskip}{0pt}}
\usepackage{bookmark}
\IfFileExists{xurl.sty}{\usepackage{xurl}}{} 
\hypersetup{
  pdftitle={Exposure-Normalized Bed and Chair Fall Rates via Continuous AI Monitoring},
  pdfauthor={Paolo Gabriel; Peter Rehani; Zack Drumm; Tyler Troy; Tiffany Wyatt; Narinder Singh},
  pdflang={en-US},
  colorlinks=true,
  linkcolor={blue},
  filecolor={Maroon},
  citecolor={Blue},
  urlcolor={blue},
  pdfcreator={LaTeX via pandoc}}

\title{Exposure-Normalized Bed and Chair Fall Rates via Continuous AI Monitoring}
\author{Paolo Gabriel \and Peter Rehani \and Zack Drumm \and Tyler Troy \and Tiffany Wyatt \and Narinder Singh}
\date{}

\begin{document}
\maketitle

\setstretch{1.15}
\section{Abstract}\label{abstract}

This retrospective cohort study used continuous AI monitoring to estimate fall rates by exposure time rather than occupied bed-days. From August 2024 to December 2025, 3,980 eligible monitoring units contributed 292,914 hourly rows, yielding probability-weighted rates of 17.8 falls per 1,000 chair exposure-hours and 4.3 per 1,000 bed exposure-hours. Within the study window, 43 adjudicated falls matched the monitoring pipeline, and 40 linked to eligible exposure hours for the primary Poisson model, producing an adjusted chair-versus-bed rate ratio of 2.35 (95\% confidence interval 0.87 to 6.33; p=0.0907). In a separate broader observation cohort (n=32 deduplicated events), 6 of 7 direct chair falls involved footrest-positioning failures. Because this was an observational study in a single health system, these findings remain hypothesis-generating and support testing safer chair setups rather than using chairs less.

\section{Introduction}\label{introduction}

Inpatient falls represent one of the most prevalent and costly preventable adverse events in acute care. Contemporary U.S. and international reports continue to show substantial event burden and injury risk across inpatient settings.\textsuperscript{1,2} Falls prevention remains a national accreditation and patient-safety priority, with The Joint Commission and national safety organizations continuing to emphasize this domain as a persistent quality challenge.\textsuperscript{3--5} Global guidance similarly identifies hospital falls prevention as an ongoing systems-level priority.\textsuperscript{6}

A central measurement gap in the inpatient falls literature is denominator choice. Most hospital fall metrics are reported per 1,000 occupied bed-days, a denominator that merges bed time, chair time, transfers, and room movement into a single exposure bucket.\textsuperscript{7} When a patient falls from a chair, the event is still counted against a bed-day denominator, which can mask position-specific hazard and dilute clinically actionable signal about chair setup and supervision.\textsuperscript{8}

Continuous AI-based patient monitoring systems create an opportunity to replace that blended denominator with position-specific exposure time. By assigning probabilistic chair and bed positions at sub-minute resolution, these systems can estimate how many falls occur per hour of chair-seated time versus per hour of bed-bound time within the same monitored population.\textsuperscript{9} This denominator-first approach aligns with standard rate modeling and supports adjusted comparisons that routine incident-reporting systems cannot produce.

This retrospective cohort analysis uses continuous AI monitoring data to estimate position-specific exposure-normalized fall rates in chair and bed contexts and then test whether the observed descriptive separation (17.8 vs 4.3 falls per 1,000 exposure-hours) persists after adjustment in a Poisson rate model. We also summarize a broader observation cohort for mechanism coding because denominator-linked chair events are sparse and mechanism hypotheses require a wider descriptive feed.

The aim is not to argue for less chair use or to promote bed confinement, given the recognized harms of low mobility in hospitalized older adults. Rather, the aim is to determine whether chair-positioned time concentrates fall opportunity enough to justify safer chair setup, clearer transfer workflows, and prospective denominator-aware monitoring studies that integrate mobility promotion with targeted fall prevention.

\section{Methods}\label{methods}

\subsection{Study Design and Setting}\label{study-design-and-setting}

This was a retrospective observational cohort study conducted using continuous AI monitoring data collected from a single regional health system and analyzed at the division level. The study period spanned August 2024 through December 2025. The study was conducted under an existing data use agreement between LookDeep Health and the health system, and constitutes a secondary analysis of de-identified monitoring records generated during routine clinical operations. No new data collection or patient contact occurred. For more information, see the ``Ethical Considerations'' section.

\subsection{Data Source}\label{data-source}

The AI monitoring platform continuously processes room-level video feeds and assigns probabilistic position estimates for each monitored patient.\textsuperscript{9} Within participating hospitals, the platform functioned as a virtual observer workflow for selected high fall-risk patients under routine operations.\textsuperscript{10,11} Per-hour position fractions (``pct\_chair,'' ``pct\_bed,'' ``pct\_ambulatory'') were extracted from the hourly monitoring cache using a validated study-window filter and percentage-sum constraint. The extraction yielded 356,391 hourly rows with a valid position-percentage sum of 100\% across all rows. Fall events were identified from AI-detected alarm records and linked to the hourly exposure structure via monitor and date-hour keys.

\subsection{Cohort Definition and Eligibility}\label{cohort-definition-and-eligibility}

A monitoring unit was defined as a unique monitor for the purposes of cohort membership. Units were classified as intervention-type if the monitor appeared in the extracted fall-event source during the run and as control-type if it appeared in the hourly exposure source without a linked extracted fall event. This outcome-defined partition was derived deterministically from the extracted \texttt{fall\_events\_source} and \texttt{hourly\_location\_aggregation} monitor sets (5,531 monitors present in hourly exposure data). Eligibility gates were applied uniformly: a unit required at least 4 observed monitor-hours during the study period (\texttt{min\_observed\_hours=4}) and a position-coverage ratio of at least 0.95 (\texttt{min\_coverage\_ratio=0.95}), defined as the proportion of eligible hours with a valid position assignment. Of 5,531 units in the hourly data, 3,980 passed both eligibility criteria (intervention: 42 eligible, 5 ineligible; control: 3,938 eligible, 1,546 ineligible). High fall-risk context in participating hospitals was informed by standard clinical screening workflows, including use of the Hendrich II model and related validation updates.\textsuperscript{12,13} The primary analysis used only eligible units.

\subsection{Exposure Measurement}\label{exposure-measurement}

Patient exposure was expressed in fractional person-hours by position type. For each eligible hourly row, the chair exposure contribution equaled \texttt{pct\_chair\ /\ 100} hours and the bed exposure contribution equaled \texttt{pct\_bed\ /\ 100} hours. Total exposure was aggregated across all intervention-eligible unit-hours in the analysis scope (not restricted to fall-positive units), yielding 320.51 chair-hours and 5,121.42 bed-hours. The analysis base comprised 292,914 rows after eligibility filtering.

\subsection{Fall Ascertainment}\label{fall-ascertainment}

Within the study window (August 2024-December 2025), 43 fall events were identified from the adjudicated consensus record and matched to the monitoring pipeline; 40 linked to eligible analysis-base hours and entered the primary exposure-normalized analysis. A broader monitoring feed (2022-2026, n=91 deduplicated events) was retained for descriptive hard-label distributions and site-summary context only. A ``hard label'' was assigned to each event based on the patient's detected position at the time of the alarm: chair, bed, room (general room space), or no\_patient (position undetected at alarm time). Hard labels were used for descriptive counts and transparency checks. For adjusted modeling, each eligible event contributed probabilistic chair/bed mass from pre-fall probabilities within the linked eligible unit-hour stratum. Of the 40 inferential events, 28 had hard labels in \{chair, bed\}; the modeled primary effective event total remained 40.0 after probabilistic redistribution across chair/bed arms, with 12 modeled events arising from room/no\_patient probability mass.

\subsection{Auxiliary Annotation Cohorts}\label{auxiliary-annotation-cohorts}

Four non-interchangeable cohorts support different descriptive tasks in this study. The study-window adjudicated cohort consisted of 43 consensus-matched events inside the August 2024-December 2025 study window. The inferential event base consisted of the 40 eligible linked fall events used in the adjusted chair-vs-bed modeling. A broader observation cohort was assembled from 37 source rows, yielding 32 included fall rows and 32 deduplicated events for descriptive mechanism coding only, after excluding 4 no\_fall rows and 1 unclassifiable empty event. A departure-aware subset of the 32 deduplicated events was then used for furniture-origin chain summaries, post-departure latency, and label-evaluation diagnostics because those analyses required benchmark linkage and sequence-level evaluability. The separate broader monitoring feed (2022-2026, n=91 deduplicated events) supports raw hard-label and site-summary descriptions only. Only the 40-event inferential base contributes to the primary RR estimate.

\begin{figure}
\centering
\includegraphics[width=0.95\linewidth,height=\textheight,keepaspectratio,alt={Figure 1. STROBE cohort flow for the denominator cohort, inferential event base, and descriptive cohorts.}]{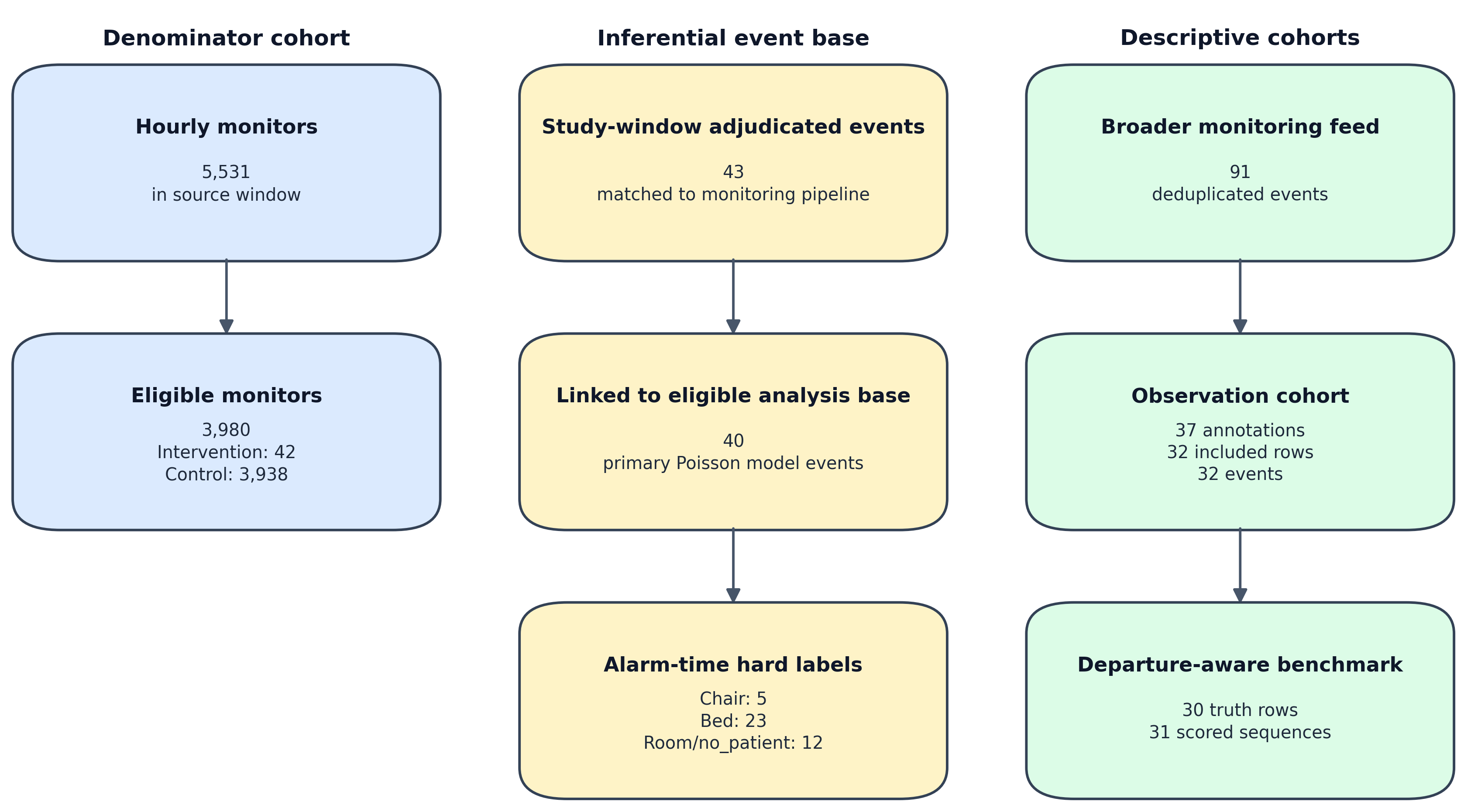}
\caption{Figure 1. STROBE cohort flow for the denominator cohort, inferential event base, and descriptive cohorts.}
\end{figure}

\subsection{Statistical Analysis}\label{statistical-analysis}

Unadjusted fall rates were calculated as events per 1,000 exposure-hours by position, extending standard falls-per-denominator conventions to position-specific hourly exposure.\textsuperscript{14} For inferential analyses, we modeled fall rates using a Poisson generalized linear model (GLM) with a log link and an offset term log(exposure\_hours), treating exposure hours as the rate denominator. The outcome for each position-unit-time stratum was the count of proportionally allocated fall events, allowing fractional counts when a single fall's risk time spanned multiple position strata; under this formulation, the Poisson GLM was used as a log-linear working mean model for rate estimation (quasi-likelihood interpretation) rather than as a strict integer-count sampling distribution.

The primary covariate of interest was patient position (chair vs.~bed; reference category = bed). Covariates included seven fixed local clock-time windows (00:00-05:59, 06:00-08:59, 09:00-11:59, 12:00-14:59, 15:00-17:59, 18:00-20:59, 21:00-23:59), day of week, calendar quarter, and division identifier, chosen a priori based on literature describing time-varying and unit-level variation in inpatient fall risk.\textsuperscript{15} The adjusted rate ratio (RR) for chair versus bed was obtained as exp(beta\_position=chair), where beta\_position=chair is the estimated coefficient for the chair indicator in the log-rate model. Ninety-five percent confidence intervals (CIs) were constructed on the log-RR scale using Wald intervals and then exponentiated back to the RR scale.

We assessed overdispersion by computing the ratio of deviance to residual degrees of freedom and the ratio of Pearson chi-squared to residual degrees of freedom; values materially greater than 1 were interpreted as evidence of extra-Poisson variation. Pre-specified inferential sufficiency checks included: (1) overdispersion ratios less than or equal to 1.5 in the primary Poisson GLM, (2) successful model convergence, and (3) minimum effective information, operationalized as at least 5 total fall events contributing to each level of the position factor in the adjusted model. When these checks passed, inferences were based on the Poisson GLM with robust sandwich covariance estimates; specifically, we computed heteroskedasticity-consistent (HC3-type) standard errors and, in sensitivity analyses, division-level clustered standard errors using cluster-robust covariance.

If overdispersion exceeded the pre-specified threshold (deviance/df or Pearson chi-squared/df greater than 1.5), we fit a negative binomial regression model with the same mean structure and offset, allowing the variance to increase quadratically with the mean as an explicit model for extra-Poisson variation. In those cases, we compared RRs and CIs from the Poisson-with-robust-SE and negative binomial models to assess robustness. Adjusted RRs were reported only when the corresponding model met all pre-specified convergence and sufficiency criteria; otherwise, adjusted position effects were labeled not estimable and not interpreted. All analyses were conducted in Python using the \texttt{statsmodels} library for GLM and negative binomial regression, with robust and clustered covariance estimators obtained via the built-in sandwich covariance options.

\subsection{Sensitivity Analyses}\label{sensitivity-analyses}

Sensitivity analyses were conducted to test time-window dependence, confidence filtering, label uncertainty, and denominator eligibility thresholds:

\begin{enumerate}
\def\labelenumi{\arabic{enumi}.}
\tightlist
\item
  Time window 00:00-05:59: Restricted analysis to overnight hours.
\item
  Time window 06:00-08:59: Restricted analysis to early-morning care hours.
\item
  Time window 09:00-11:59: Restricted analysis to late-morning hours.
\item
  Time window 12:00-14:59: Restricted analysis to midday hours.
\item
  Time window 15:00-17:59: Restricted analysis to late-afternoon hours.
\item
  Time window 18:00-20:59: Restricted analysis to evening hours.
\item
  Time window 21:00-23:59: Restricted analysis to late-evening hours.
\item
  Weekday-only: Restricted to Monday-Friday; flagged for insufficient data.
\item
  Weekend-only: Restricted to Saturday-Sunday; flagged for insufficient data.
\item
  Position-certainty hours only: Restricted to rows where either chair or bed exposure fraction was at least 0.50 in the indexed hour.
\item
  Misclassification scenarios: Symmetric 10\%, 20\%, 30\% chair/bed swaps and one-sided 20\% swaps (chair-to-bed and bed-to-chair).
\item
  Eligibility threshold sensitivity: Recomputed at \texttt{min\_observed\_hours} values of 4, 12, 24, and 48 hours.
\item
  Furniture-origin reclassification: Room/no\_patient falls that originated from chairs (identified by the \texttt{last\_furniture} field in the consensus observation cohort) were reclassified as chair-associated events. For matched hours in the analysis base, fall attribution was overridden to 100\% chair exposure. This tests whether the chair risk signal holds when chair-origin room falls are included.
\end{enumerate}

\subsection{Furniture-Exit Detection Concordance}\label{furniture-exit-detection-concordance}

To validate the AI pipeline's ability to detect when patients leave furniture positions, we compared two AI-derived departure signals against consensus-annotated \texttt{furniture\_departure\_time} for events with known last-furniture provenance. The first signal was the nudge boundary crossing: the first frame-level transition from green to yellow or red in the operational \texttt{nudge\_state} risk indicator, representing the system's real-time alert trigger. The second was the location-label transition: the first change in \texttt{dominant\_location\_label} away from the patient's known last-furniture position. For each consensus event with a valid departure time, a +/-5-minute search window was applied to the second-level panel data, and concordance was quantified as bias (AI minus consensus, in seconds), mean absolute error (MAE), and percentile-based absolute error distributions.

\subsection{Label Evaluation}\label{label-evaluation}

Automated label quality was evaluated against the departure-aware benchmark subset of the v3 observation cohort (30 truth rows spanning 31 scored sequences). Metrics are reported as descriptive evidence of label uncertainty: macro F1, expected calibration error (ECE, 10-bin), detection F1 (event-level), detection precision, detection recall, and latency mean absolute error (MAE). The observed metrics remain a material limitation and motivate sensitivity analyses and conservative interpretation.

\subsection{Ancillary Multimodal LLM Benchmark}\label{ancillary-multimodal-llm-benchmark}

Three multimodal LLMs (Gemini 2.5 Flash, Gemini 2.5 Pro, Gemini 3.1 Pro Preview) independently processed de-identified adjudicated event clips (falls and non-falls) from the v2 consensus evaluation package (81 clips; distinct from the 32-event mechanism observation cohort). Each model classified fall detection, pre-fall location, last-furniture provenance, mechanism tags, and fall timing. The primary comparison view was a 3-way overlap subset (n=42 visible primary rows, 38 visible fall events, 4 visible non-fall events) where all three models returned a successful analysis. Metrics reported include fall sensitivity, non-fall specificity, pre-fall location accuracy (detected visible falls), and fall-time MAE (detected visible falls). This benchmark uses a different denominator structure than the primary livestream pipeline and is reported as an ancillary validation, not a head-to-head comparison.

\subsection{Interpretation Guardrails}\label{interpretation-guardrails}

This project uses pre-specified internal gates and reporting rules to prevent over-interpretation. In this manuscript, ``descriptive'' refers to reporting cohort characteristics, exposure-normalized rates, and label-evaluation metrics. ``Inferential'' refers to reporting adjusted chair-vs-bed rate ratios with uncertainty only when pre-specified data sufficiency criteria are met. Gate labels indicate internal process checks (data extraction, label evaluation, and de-identification) and do not imply statistical significance or causal validity.

\subsection{Mechanism Taxonomy}\label{mechanism-taxonomy}

A separate broader observation cohort of 32 deduplicated fall events (from 32 included fall rows across 37 source rows, with 4 no\_fall rows and 1 unclassifiable empty event excluded) was assembled for mechanism coding via a structured dual-review process. Each event was classified into one of three mechanism categories: footrest/positioning, transfer failure, or other/unclassified. Among the 7 direct chair falls in this observation cohort, 6 carried a footrest/positioning tag. This cohort is not a denominator-linked study-window cohort and is reported for descriptive mechanism characterization only.

\subsection{Ethics and Data Governance}\label{ethics-and-data-governance}

The data used in this retrospective study were collected from patients admitted to one of eleven hospital partners across three different states in the USA. The study and handling of data followed the guidelines provided by CHAI standards. Access to these data was granted to the researchers through a Business Associate Agreement (BAA) specifically for monitoring patients at high risk of falls. In compliance with the Health Insurance Portability and Accountability Act (HIPAA), patients provided written informed consent for monitoring as part of their standard inpatient care. To ensure patient privacy, all video data were blurred prior to storage, and no identifiable information is included in this work. Face-blurred frames were used only for training purposes. Faces were manually labeled on fully blurred images, and the raw images were then treated with a local Gaussian blur in the facial regions, ensuring privacy without compromising model training quality. The outcomes of this analysis did not influence patient care or clinical outcomes.

\section{Results}\label{results}

Using position-specific exposure hours rather than occupied bed-days as the denominator, chair time showed a higher descriptive fall rate than bed time (17.8 vs 4.3 falls per 1,000 exposure-hours). Within the study window, 43 adjudicated events matched the monitoring pipeline and 40 linked to eligible analysis-base hours for adjusted modeling, yielding a primary chair-vs-bed RR of 2.35 (95\% CI 0.87 to 6.33; p=0.0907). In a separate broader observation cohort (n=32), 6 of 7 direct chair falls carried a footrest/positioning tag.

\textbf{Table 1. Cohort summary statistics for the chair-falls retrospective cohort analysis.}

{\def\LTcaptype{none} 
\begin{longtable}[]{@{}
  >{\raggedright\arraybackslash}p{(\linewidth - 2\tabcolsep) * \real{0.4918}}
  >{\raggedright\arraybackslash}p{(\linewidth - 2\tabcolsep) * \real{0.4918}}@{}}
\toprule\noalign{}
\endhead
\bottomrule\noalign{}
\endlastfoot
\textbf{Parameter} & \textbf{Value} \\
Health system & {[}Regional Health System{]} (analyzed at division level) \\
Study period & August 2024 - December 2025 \\
Total monitors (hourly data) & 5,531 \\
Total monitors (cohort map) & 5,570 \\
Eligible monitors (both gates) & 3,980 / 5,531 (72.0\%) \\
Intervention-eligible & 42 \\
Control-eligible & 3,938 \\
Ineligible monitors & 1,551 \\
Intervention-ineligible & 5 \\
Control-ineligible & 1,546 \\
Total hourly exposure rows (valid) & 356,391 \\
Analysis base rows (after eligibility) & 292,914 \\
Study-window adjudicated events matched to pipeline & 43 \\
Inferential fall events linked to analysis base & 40 \\
Broader monitoring feed (2022-2026, descriptive only) & 91 \\
Broader observation cohort (descriptive mechanism coding) & 32 deduplicated events \\
Departure-aware benchmark subset & 30 truth rows / 31 scored sequences \\
Eligibility gates applied & min\_observed\_hours=4; min\_coverage\_ratio=0.95 \\
Primary model & Poisson GLM; offset=log(exposure\_hours) \\
\end{longtable}
}

\subsection{Descriptive Fall Rates by Position (Unadjusted)}\label{descriptive-fall-rates-by-position-unadjusted}

Among intervention-eligible units, the probability-weighted descriptive chair fall rate was 17.8 per 1,000 exposure-hours (5.69 expected falls / 320.51 chair-exposure-hours) and the corresponding bed rate was 4.3 per 1,000 exposure-hours (22.05 expected falls / 5,121.42 bed-exposure-hours). Hard-label rates are shown alongside for transparency (chair 15.6; bed 4.5 per 1,000 exposure-hours), but the probability-weighted rates better match the event allocation used in adjusted modeling.

\begin{figure}
\centering
\includegraphics[width=0.85\linewidth,height=\textheight,keepaspectratio,alt={Figure 2. Unadjusted fall rates per 1,000 exposure-hours by patient position among intervention-eligible monitoring units.}]{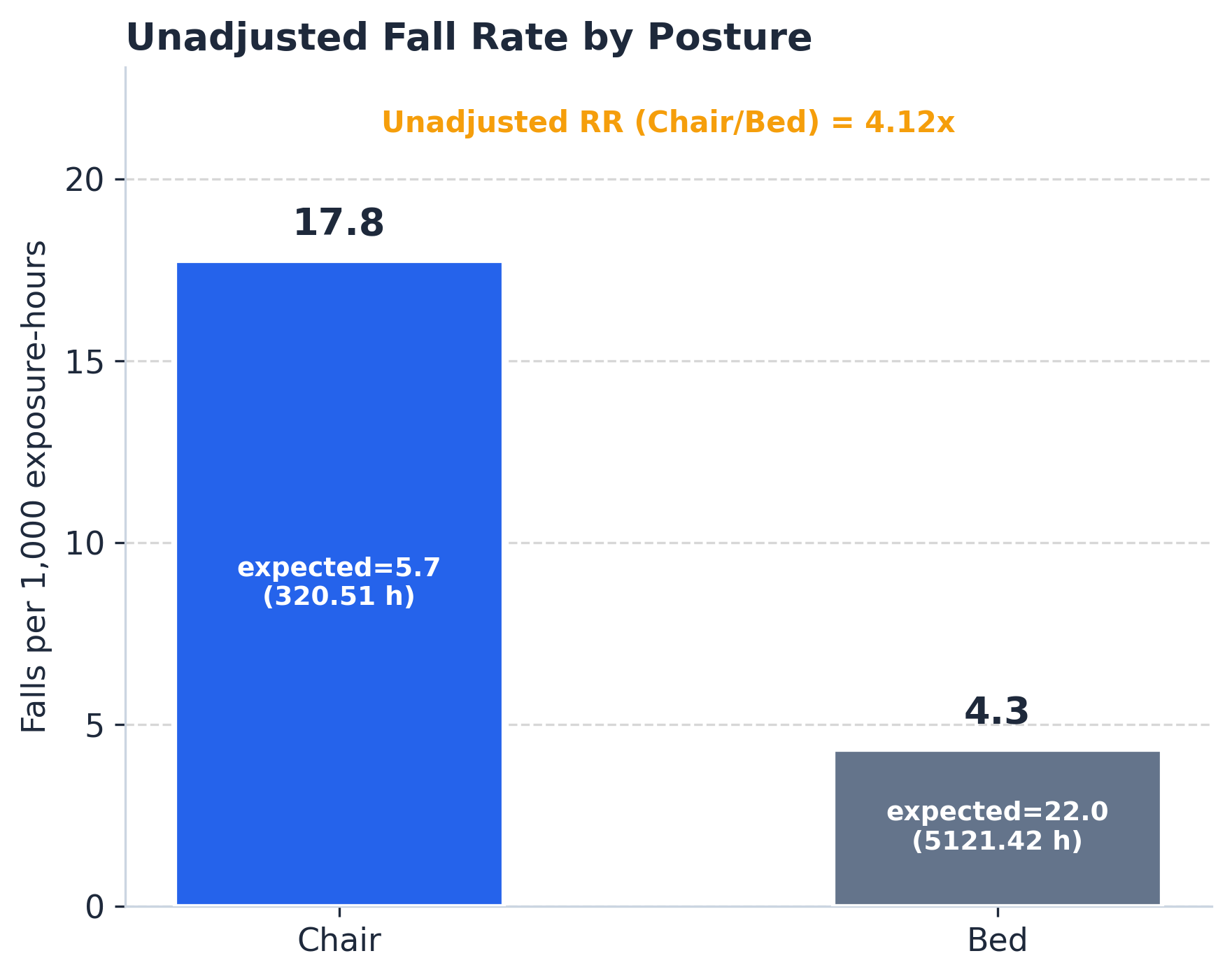}
\caption{Figure 2. Unadjusted fall rates per 1,000 exposure-hours by patient position among intervention-eligible monitoring units.}
\end{figure}

\textbf{Table 2. Unadjusted fall rates by patient position, intervention-eligible units.}

{\def\LTcaptype{none} 
\begin{longtable}[]{@{}
  >{\raggedright\arraybackslash}p{(\linewidth - 8\tabcolsep) * \real{0.1172}}
  >{\raggedright\arraybackslash}p{(\linewidth - 8\tabcolsep) * \real{0.2266}}
  >{\raggedright\arraybackslash}p{(\linewidth - 8\tabcolsep) * \real{0.1797}}
  >{\raggedright\arraybackslash}p{(\linewidth - 8\tabcolsep) * \real{0.1797}}
  >{\raggedright\arraybackslash}p{(\linewidth - 8\tabcolsep) * \real{0.2812}}@{}}
\toprule\noalign{}
\endhead
\bottomrule\noalign{}
\endlastfoot
\textbf{Position} & \textbf{Hard-label fall events} & \textbf{Exposure (hours)} & \textbf{Rate per 1,000 h} & \textbf{Expected rate (pre-specified)} \\
Chair & 5 & 320.51 & 15.6 & 17.8 \\
Bed & 23 & 5,121.42 & 4.5 & 4.3 \\
\end{longtable}
}

\begin{quote}
Note: Expected rates are probability-weighted descriptive rates derived from pre-fall location probabilities and align with the adjusted-model event allocation. Hard-label counts reflect AI-assigned position at event time. Analysis restricted to intervention-eligible units only (n=42 intervention monitors in analysis base).
\end{quote}

\subsection{Primary Adjusted Relative Risk}\label{primary-adjusted-relative-risk}

The adjusted Poisson GLM specifications executed on the 40-event inferential base. The primary adjusted chair-vs-bed RR was 2.35 (95\% CI 0.87 to 6.33; p=0.0907). HC3 standard errors preserved direction with a similar interval width (RR 2.35, 95\% CI 0.93 to 5.94; p=0.0709). A division-clustered sensitivity fit yielded the same point estimate with a narrower interval (RR 2.35, 95\% CI 1.89 to 2.92; p\textless0.0001), but this row is exploratory only because it is based on nine intervention divisions within one health system and the model showed underdispersion rather than overdispersion (deviance/df = 0.13; Pearson chi-squared/df = 0.49). The position-certainty-only restriction was non-estimable, and the pre-specified negative-binomial fallback was not triggered.

\subsection{Sensitivity Analyses}\label{sensitivity-analyses-1}

\textbf{Table 3. Adjusted rate ratios and sensitivity analyses.}

{\def\LTcaptype{none} 
\begin{longtable}[]{@{}
  >{\raggedright\arraybackslash}p{(\linewidth - 10\tabcolsep) * \real{0.2603}}
  >{\raggedright\arraybackslash}p{(\linewidth - 10\tabcolsep) * \real{0.0616}}
  >{\raggedright\arraybackslash}p{(\linewidth - 10\tabcolsep) * \real{0.1027}}
  >{\raggedright\arraybackslash}p{(\linewidth - 10\tabcolsep) * \real{0.0959}}
  >{\raggedright\arraybackslash}p{(\linewidth - 10\tabcolsep) * \real{0.1164}}
  >{\raggedright\arraybackslash}p{(\linewidth - 10\tabcolsep) * \real{0.3493}}@{}}
\toprule\noalign{}
\endhead
\bottomrule\noalign{}
\endlastfoot
\textbf{Analysis} & \textbf{RR} & \textbf{95\% CI} & \textbf{p-value} & \textbf{Events (n)} & \textbf{Notes} \\
Primary (all hours) & 2.35 & 0.87 to 6.33 & 0.0907 & 40 & estimable \\
Primary (HC3) & 2.35 & 0.93 to 5.94 & 0.0709 & 40 & HC3 SE \\
Exploratory clustered SE (division) & 2.35 & 1.89 to 2.92 & \textless0.0001 & 40 & 9 intervention divisions; not a primary estimate \\
Position-certainty hours only & NE & NE & --- & 38 & insufficient\_data \\
Chair-origin reclassified & 2.35 & 0.87 to 6.33 & 0.0907 & 40 & furniture-origin sensitivity \\
Eligibility threshold \textgreater= 12 h & 2.39 & 0.88 to 6.47 & 0.0859 & 38 & estimable \\
Eligibility threshold \textgreater= 24 h & 2.43 & 0.89 to 6.63 & 0.0831 & 36 & estimable \\
Eligibility threshold \textgreater= 48 h & 2.42 & 0.84 to 6.98 & 0.1008 & 30 & estimable \\
\end{longtable}
}

\begin{quote}
Note: Abbreviations: RR, rate ratio; CI, confidence interval; NE, non-estimable. Non-estimable: fewer than 5 events in one or both position arms within stratum, per pre-specified inferential sufficiency criteria. The clustered row is exploratory only because it uses nine intervention divisions and sits within an underdispersed Poisson fit (deviance/df = 0.13; Pearson chi-squared/df = 0.49). All seven pre-specified local-time window restriction analyses (00:00-05:59 through 21:00-23:59) were non-estimable in this run because one or both position arms had fewer than 5 modeled events. Those rows are retained in the run artifact for completeness but omitted from Table 3 for readability. Misclassification sensitivity scenarios were evaluated in five pre-specified swaps: symmetric 10\%, 20\%, and 30\% chair/bed reassignments, plus one-sided 20\% chair-to-bed and bed-to-chair swaps. Across estimable scenarios, RR ranged from 4.49 to 13.17; the one-sided 20\% chair-to-bed scenario became non-estimable, underscoring how strongly effect magnitude depends on position-label assumptions. These are scenario-based results and should not be interpreted as stable point estimates.
\end{quote}

\begin{figure}
\centering
\includegraphics[width=0.92\linewidth,height=\textheight,keepaspectratio,alt={Figure 3. Forest plot of adjusted rate ratios across the primary and sensitivity analyses.}]{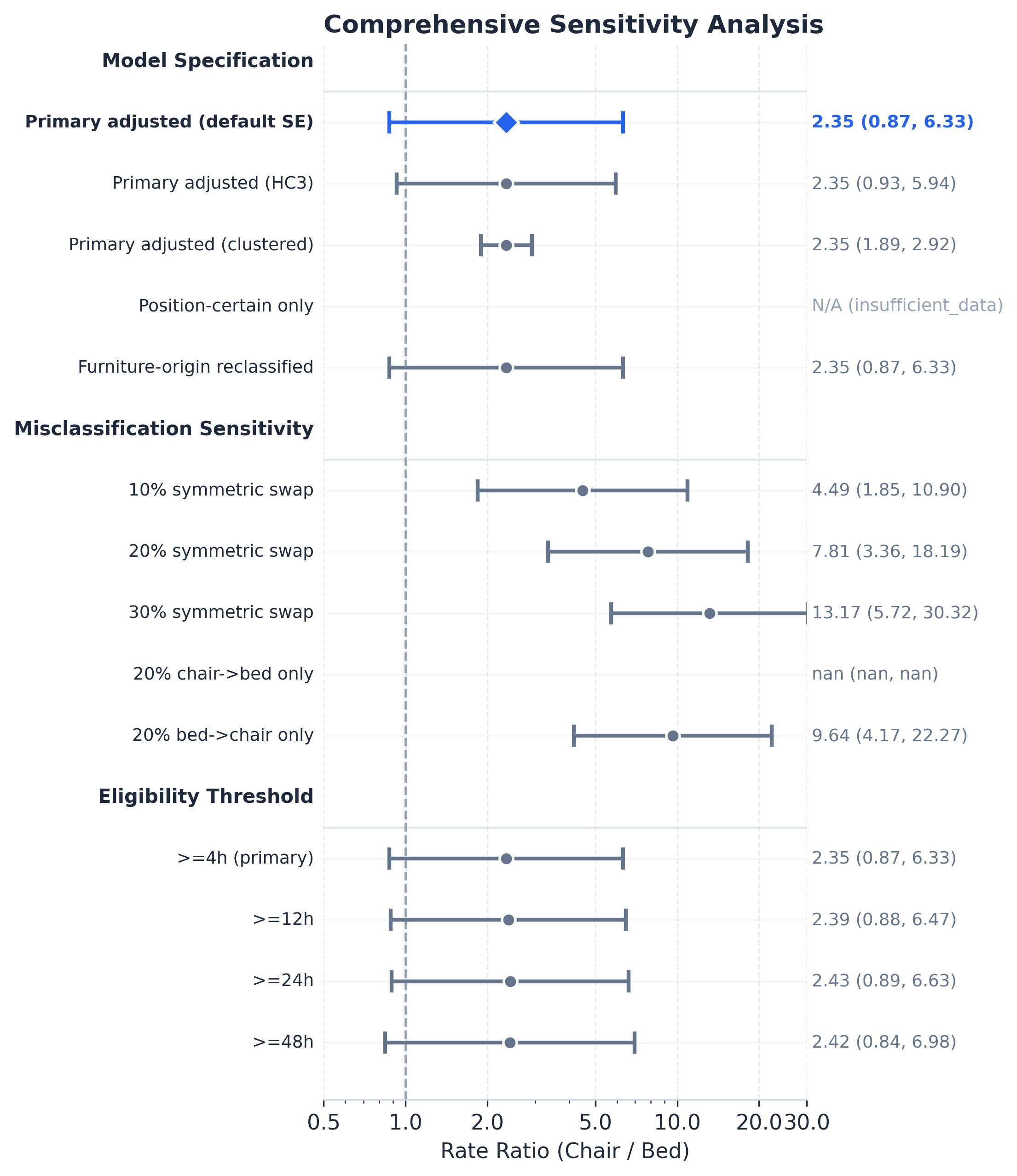}
\caption{Figure 3. Forest plot of adjusted rate ratios across the primary and sensitivity analyses.}
\end{figure}

\subsection{Mechanism Taxonomy (Observation Cohort)}\label{mechanism-taxonomy-observation-cohort}

The mechanism taxonomy was developed from a broader observation cohort of 32 deduplicated fall events (from 32 included fall rows) dual-reviewed by clinical observers. One event with no annotation data was excluded as unclassifiable. This cohort was assembled for clinical QA and mechanism characterization; it is not denominator-linked, it is not limited to the study window, and the counts below should not be interpreted as base rates in the inferential population. For room/no\_patient context preservation, provenance modifiers were retained from the consensus annotation layer. In deduplicated mechanism events, last-furniture provenance was populated in 13/13 room events, with a breakdown of bed (10) and chair (3). Where both departure and fall clocks were available, time-since-departure had median 0.30 minutes (IQR 0.12-1.21; max 2.50; n=13).

\begin{figure}
\centering
\includegraphics[width=0.92\linewidth,height=\textheight,keepaspectratio,alt={Figure 4. Fall mechanism taxonomy stratified by pre-fall patient position (observation cohort, n=32 events).}]{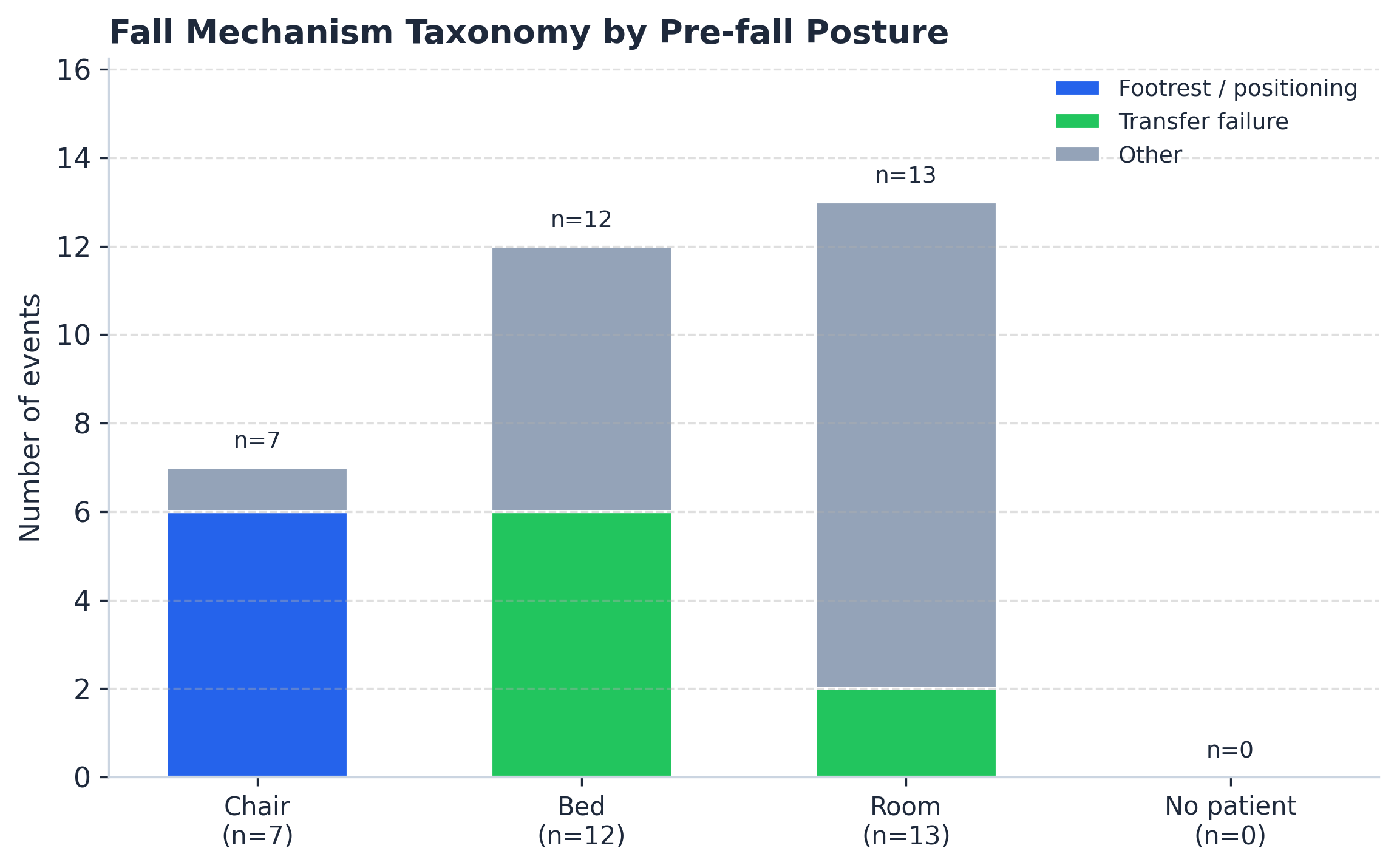}
\caption{Figure 4. Fall mechanism taxonomy stratified by pre-fall patient position (observation cohort, n=32 events).}
\end{figure}

Representative de-identified stills from the broader observation review illustrate the three fall contexts used during qualitative mechanism review: floor falls, bed falls, and chair falls. These frames are included as visual examples rather than denominator-linked evidence and do not alter the quantitative taxonomy counts reported above.

\begin{figure}
\centering
\includegraphics[width=0.95\linewidth,height=\textheight,keepaspectratio,alt={Figure 5. Representative de-identified video stills illustrating floor, bed, and chair fall sequences. Columns show the pre-fall position, fall event, and post-fall response; faces are blurred for privacy.}]{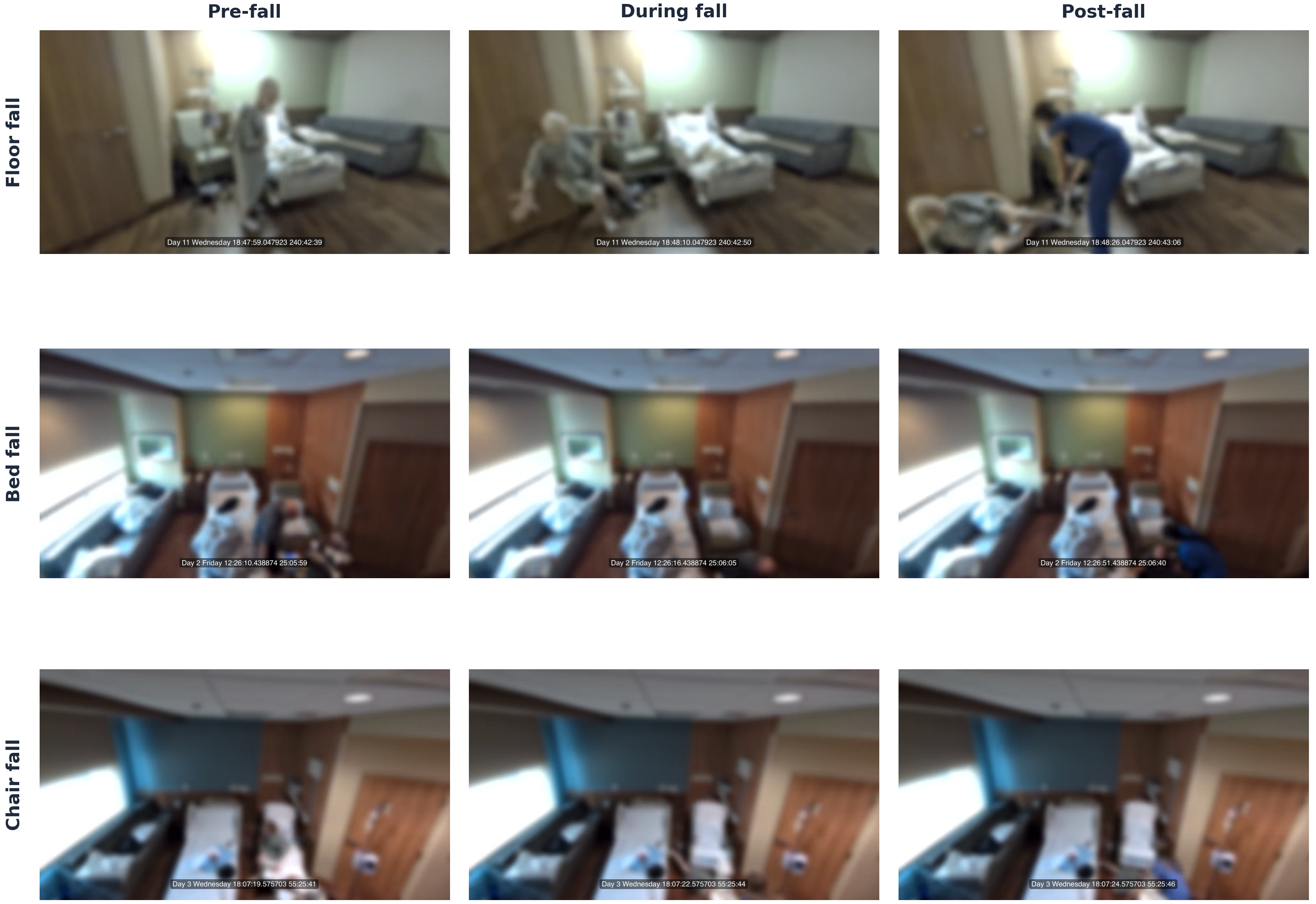}
\caption{Figure 5. Representative de-identified video stills illustrating floor, bed, and chair fall sequences. Columns show the pre-fall position, fall event, and post-fall response; faces are blurred for privacy.}
\end{figure}

Detailed mechanism, furniture-origin, and ancillary model tables are reported in Appendix Tables A1-A3.

\subsection{Furniture-Origin Chain Analysis}\label{furniture-origin-chain-analysis}

Among the 32 deduplicated fall events with evaluable furniture-origin chain annotations, classification identified 7 direct chair falls, 12 direct bed falls, 3 chair-origin room falls (events where the patient was last observed in a chair before falling in the room), and 10 bed-origin room falls. Including the 3 chair-origin room events with the 7 direct chair falls yields 10 chair-associated events, compared to 7 when using only direct chair classification.

\subsection{Post-Departure Latency}\label{post-departure-latency}

For events with both a known furniture departure time and a fall time, the post-departure latency quantifies the interval between leaving furniture and falling. Among bed-origin events, the median latency was 18 seconds (IQR 6-69; max 136; n=10). Among chair-origin events, the median latency was 50 seconds (max 150; n=3). These latencies characterize a clinically actionable post-departure risk window during which patients are transitioning and may benefit from enhanced monitoring or staff proximity.

\subsection{Furniture-Origin Reclassification Sensitivity}\label{furniture-origin-reclassification-sensitivity}

When the 3 chair-origin room falls were assessed for chair-origin reclassification in the Poisson GLM, the adjusted chair-vs-bed RR remained 2.35 (95\% CI 0.87 to 6.33; p=0.0907; n=40 events). In this run, the chair-origin room review did not materially change modeled chair/bed event allocation within the eligible analysis base. This reclassification sensitivity result is reported alongside the primary and other sensitivity analyses in Table 3.

\subsection{Furniture-Exit Detection Concordance}\label{furniture-exit-detection-concordance-1}

Among consensus events with valid furniture departure timestamps, two AI departure signals were evaluated against the consensus reference. In this run, no events met evaluability criteria for either signal after linkage and filtering, so bias and MAE estimates were not estimable. This indicates a current instrumentation gap rather than evidence of zero timing error and should be treated as a data-availability limitation for exit-concordance validation.

\subsection{Hard-Label Position Distribution at Alarm (Broader Monitoring Feed)}\label{hard-label-position-distribution-at-alarm-broader-monitoring-feed}

Within the broader monitoring feed (2022-2026, n=91 deduplicated events), the hard-label position distribution was: bed (n=48, 52.7\%), no\_patient (n=23, 25.3\%), room (n=12, 13.2\%), and chair (n=8, 8.8\%). The no\_patient category reflects events where the model did not assign a confident position label at alarm time. Hard labels are used for descriptive counts; adjusted modeling uses probabilistic allocation inside the 40-event inferential base.

\subsection{Label Evaluation Metrics (Descriptive)}\label{label-evaluation-metrics-descriptive}

Automated label quality was assessed against the v3 adjudicated benchmark subset (30 truth rows, 31 scored sequences). Metrics are reported as descriptive evidence of label uncertainty: macro F1 = 0.528, expected calibration error (10-bin ECE) = 0.450, detection F1 = 0.846, detection precision = 1.000, detection recall = 0.733, latency MAE = 37.9 seconds (p50 = 22s, p90 = 83s). The perfect detection precision indicates that the model generates no false positive fall alarms, but recall of 0.733 means it still misses roughly one in four fall events at the detection level. The macro F1 of 0.528 reflects persistent multi-class position-classification limitations, especially away from the bed class. In a confidence-filtered descriptive sensitivity analysis, rates remained close to the all-event estimates (chair 17.1 vs 17.8; bed 4.2 vs 4.3 per 1,000 exposure-hours), providing only partial reassurance in the presence of poor calibration.

\subsection{Ancillary Multimodal LLM Benchmark}\label{ancillary-multimodal-llm-benchmark-1}

The three models exhibited a sensitivity-specificity tradeoff consistent with the primary pipeline's label-quality findings. Gemini 2.5 Flash achieved the highest fall sensitivity (0.842) but the lowest specificity (0.750) and poorest timing accuracy. Gemini 3.1 Pro Preview achieved perfect specificity (1.000) and the best timing accuracy (MAE 106s) but missed over half of visible falls (sensitivity 0.447). No model achieved both high sensitivity and high location accuracy, reinforcing the persistent difficulty of automated position classification in hospital monitoring contexts. Aggregate metrics for the overlap subset are reported in Appendix Table A3.

\subsection{Falls by Hospital Site}\label{falls-by-hospital-site}

Within the broader monitoring feed (2022-2026, n=91 deduplicated events), falls were observed across 10 divisions within the health system. Site names are not reported in this manuscript per de-identification requirements. The top seven divisions accounted for 85 of 91 broader-feed events, with the largest contributing 31 events and the second largest contributing 18 events.

\section{Discussion}\label{discussion}

\subsection{Magnitude of Chair-Seated Risk}\label{magnitude-of-chair-seated-risk}

The main contribution of this analysis is denominator refinement rather than a definitive effect estimate. Once chair and bed time were expressed as separate exposure-hours, the descriptive rate contrast was substantial (17.8 vs 4.3 per 1,000 exposure-hours), and the adjusted chair-vs-bed RR remained above 1.0 at 2.35. However, the primary confidence interval crossed 1.0 (95\% CI 0.87 to 6.33; p=0.0907), so the complete experiment should be interpreted as an elevated but uncertain signal rather than a confirmed multi-fold effect. The inferential event base remains modest (43 adjudicated study-window events, 40 eligible linked events), so interpretation should remain cautious and hypothesis-generating rather than confirmatory.

\subsection{Mechanism Insights}\label{mechanism-insights}

In the broader observation cohort, 6 of 7 direct chair falls carried a footrest/positioning tag. This concentration points to a consistent upstream problem: incomplete chair setup or unsupported lower-extremity position. Footrest failure specifically may reflect rushed patient transfers, inadequate nursing training on chair ergonomics, or chair designs that are difficult to configure for patients with lower-limb weakness. Because this 32-event cohort is descriptive, broader than the study window, and not denominator-linked, the finding should be treated as signal generation for prospective testing rather than as a population prevalence estimate.

\subsection{Time-Window Sensitivity Analysis}\label{time-window-sensitivity-analysis}

Time-window stratifications remained underpowered for inferential reporting. All seven pre-specified local-time windows (00:00-05:59, 06:00-08:59, 09:00-11:59, 12:00-14:59, 15:00-17:59, 18:00-20:59, and 21:00-23:59) were non-estimable because one or both position arms fell below the pre-specified inferential sufficiency threshold. These strata should therefore be interpreted as planning signals rather than confirmatory subgroup effects.

Misclassification sensitivity analyses showed that plausible label-swap scenarios can materially move the estimated effect size. Across estimable pre-specified scenarios, RR ranged from 4.49 to 13.17, while the one-sided 20\% chair-to-bed reassignment became non-estimable. These patterns reinforce sensitivity of effect magnitude to label error assumptions.

\subsection{Ancillary LLM Benchmark}\label{ancillary-llm-benchmark}

The ancillary Gemini benchmark provides an independent model family's perspective on the same fall-detection and position-classification task. The sensitivity-specificity tradeoff observed across the three Gemini models - where higher recall came at the cost of lower specificity and worse timing - is consistent with the primary pipeline's label-quality profile and reinforces the conclusion that automated fall-position classification remains a material limitation in hospital monitoring contexts.

Conservative models (high specificity, low sensitivity) missed many visible falls; the false-negative tag analysis indicates these were disproportionately subtle or furniture-proximal events tagged with slip, support\_bed, slow, or back in the consensus annotations. This benchmark uses different denominators (3-way overlap, n=42 visible primary rows) and scoring (raw GT location vs departure-aware truth) than the primary pipeline, so direct ranking between the two is not appropriate. The findings support the paper's conclusion that prospective improvement in automated position classification is needed before label-derived metrics can support confirmatory inference.

\subsection{Latent Bias Assessment}\label{latent-bias-assessment}

Position classification ambiguity. The consensus observation dataset contains no intermediate position labels and no event-level certainty scores, so classification error cannot be directly measured from the observation cohort. The scenario-based misclassification analyses therefore remain the main quantitative guardrail against this threat. Across estimable scenarios, effect size changed materially, indicating that the magnitude of the chair-vs-bed association is sensitive to label error assumptions even when direction is preserved.

Temporal confounding. Daypart summaries show expected fall mass concentrated in waking and care-transition hours, with the highest mean chair probability in those same windows (Figure 6), and the direct-chair fall observations show a similar daytime skew. The primary model adjusts for seven local-time windows, which is a reasonable attempt to absorb this structure, but sparse chair-event counts limit how completely time-of-day clustering can be controlled. Residual temporal confounding would most likely inflate the observed chair-vs-bed rate ratio.

\begin{figure}
\centering
\includegraphics[width=0.95\linewidth,height=\textheight,keepaspectratio,alt={Figure 6. Daypart distribution of expected falls with mean chair probability across seven predefined local time windows. Bars summarize expected falls by pre-fall location, and chair bars are shaded by mean chair probability to provide temporal-confounding context rather than a direct-chair event histogram.}]{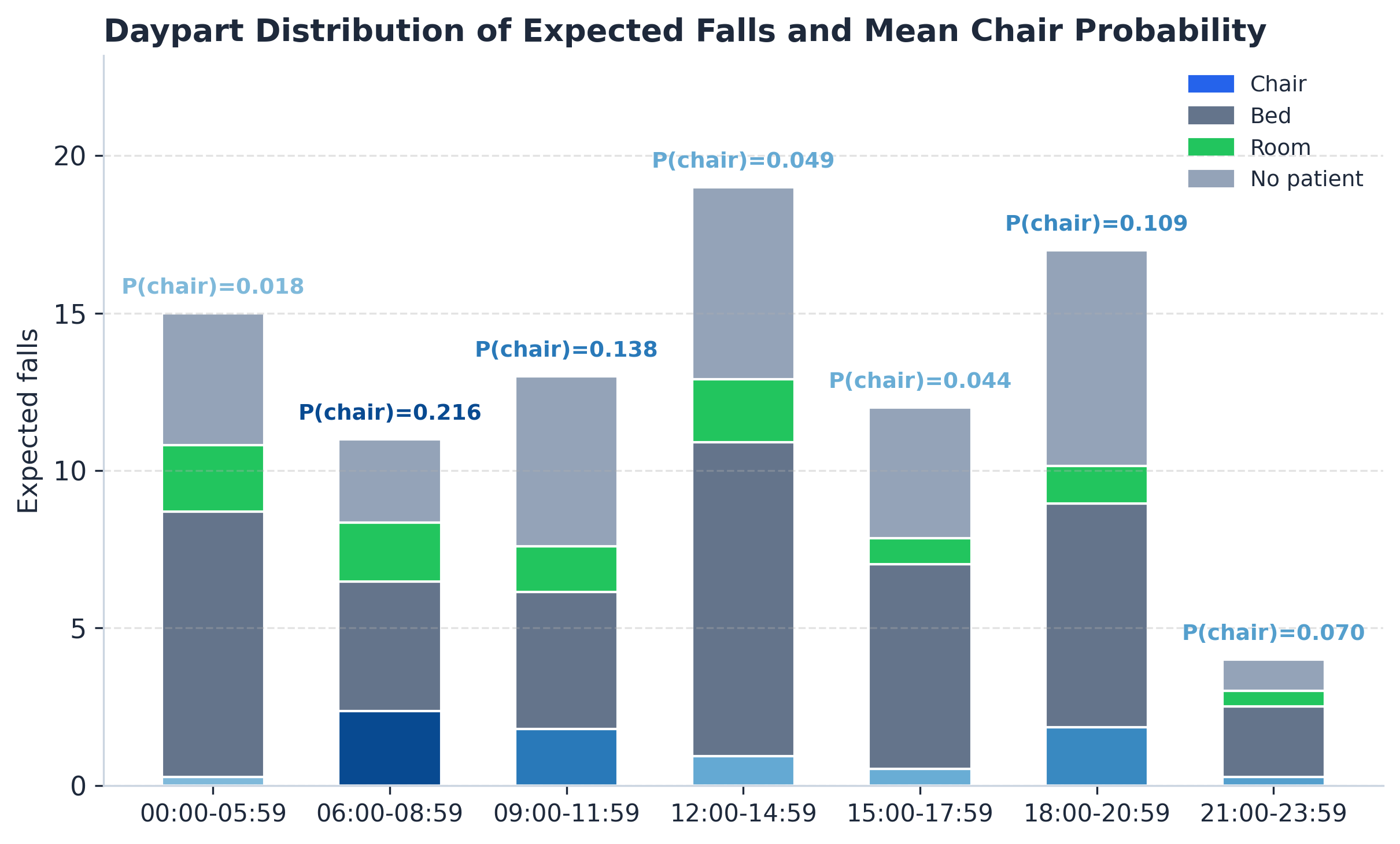}
\caption{Figure 6. Daypart distribution of expected falls with mean chair probability across seven predefined local time windows. Bars summarize expected falls by pre-fall location, and chair bars are shaded by mean chair probability to provide temporal-confounding context rather than a direct-chair event histogram.}
\end{figure}

Selection into position. The monitoring dataset contains no patient-level acuity, mobility, or staffing variables, so confounding by indication remains the most important unmeasured threat. The relative homogeneity of chair-fall mechanisms offers some support for a position-specific hazard, but it does not eliminate the possibility that patient selection into chair placement accounts for part or all of the observed association.

Bias audit summary. The observed chair-vs-bed rate ratio remains sensitive to both classification error assumptions and residual confounding. The primary interval (0.87 to 6.33) already allows for a substantially smaller effect than the point estimate suggests, and the true causal effect of chair positioning could be materially attenuated or null once temporal and acuity confounders are better measured. These findings reinforce the manuscript's framing of results as hypothesis-generating rather than confirmatory.

\subsection{Post-Chair Risk Window and Furniture-Origin Analysis}\label{post-chair-risk-window-and-furniture-origin-analysis}

The furniture-origin chain analysis suggests that chair-associated risk extends beyond direct chair falls: three room-classified events in the observation cohort were preceded by chair departure, expanding the chair-associated event count from 7 to 10. The post-departure latency findings identify a short interval between leaving furniture and falling, supporting a clinically relevant transition window for future prospective monitoring work.

\subsection{Limitations}\label{limitations}

Several important limitations must be acknowledged. First, this is an observational, single health system study; causal inference is not supported. The adjusted RR controls for measured temporal and site-level confounders but cannot address unmeasured confounding such as patient acuity, nursing staffing ratios, or patient mobility classification, none of which were available in the monitoring dataset. Directionality of this bias is uncertain, but one plausible pathway is selection into chair placement among patients with greater transfer attempts or mobility, which could inflate observed chair-associated fall risk independent of position itself. Unit-level occupancy and operational conditions are also known to correlate with falls and fall-related injuries, reinforcing residual confounding risk.\textsuperscript{16}

Second, label quality is a material constraint. The automated position classifier achieved a macro F1 of 0.528 against the adjudicated benchmark subset, with ECE = 0.450. That calibration error means the probability mass used for chair-versus-bed allocation could be biased, not merely noisy. The detection F1 of 0.846 (precision 1.000 / recall 0.733) indicates that fall detection itself is specific but incomplete. Some falls in the analysis base may be misclassified by position, and falls not detected by the model are absent from the analysis entirely, creating potential outcome ascertainment bias. Confidence-filtered descriptive rates stayed close to the all-event estimates (chair 17.1 vs 17.8; bed 4.2 vs 4.3 per 1,000 exposure-hours), which offers partial reassurance about rank ordering but does not resolve the calibration problem.

Third, the study is limited to a single regional health system, and the generalizability of both the rate estimates and the mechanism findings to other health systems, patient populations, or monitoring platforms is unknown. Transportability may be constrained by differences in monitor placement, camera fields-of-view, nursing workflows, chair hardware, and local staffing models.

Fourth, this study did not capture injury outcomes. Falls without injury are included alongside fall-with-injury events in the analysis; severity-weighted analyses were not possible with available data.

Fifth, the observation cohort used for mechanism coding (n=32 deduplicated events) is a convenience sample collected for internal QA, not a probability sample of all falls in the analysis population. Mechanism proportions should be treated as hypothesis-generating, not as population prevalence estimates.

Sixth, the furniture-origin reclassification is anchored to the observation cohort (n=32 deduplicated events) rather than to the full analysis base. Chair-origin room events not captured in that cohort cannot be reclassified, so this sensitivity analysis may understate the true number of chair-associated events. Additionally, the furniture-exit concordance is limited to events with second-level panel coverage and valid departure timestamps, which may not represent all transition events in the study period.

\subsection{Clinical Implications}\label{clinical-implications}

Chair positioning remains clinically important for mobilization, delirium prevention, and avoidance of the functional harms associated with prolonged bed rest.\textsuperscript{17} The implication of these findings is therefore safer chair use, not less chair use.

What can inform near-term practice (hypothesis-generating): The mechanism pattern supports targeted chair-setup confirmation before leaving a chair-positioned patient unattended (for example, footrest position, leg support, and call-light accessibility), framed as a QA/protocol hypothesis rather than a proven intervention. Existing implementation literature suggests that protocol reliability depends on both local workflow adoption and explicit implementation strategy selection.\textsuperscript{18--20}

What requires prospective confirmation: Stable adjusted chair-vs-bed estimates and transportability to other sites or platforms require prospective data with larger chair exposure and event counts, improved position-label calibration, and key clinical confounders (acuity, mobility status, staffing) captured. This technology pathway should be framed within nursing-led AI practice integration rather than as a stand-alone technical intervention.\textsuperscript{21}

\section{Conclusions}\label{conclusions}

Position-specific exposure denominators identified higher descriptive fall rates during chair time than bed time (17.8 vs 4.3 probability-weighted falls per 1,000 exposure-hours), and adjusted modeling estimated a chair-vs-bed RR of 2.35 (95\% CI 0.87 to 6.33). In a separate broader observation cohort, 6 of 7 direct chair falls involved footrest-positioning failures, pointing to a plausible modifiable prevention target. These findings remain hypothesis-generating and support safer chair use workflows, but they require prospective validation before causal or policy-level conclusions.

\section{Keywords}\label{keywords}

Patient Positioning; Accidental Falls; Video Recording; Hospital Units; Poisson Regression

\section{Conflict of Interest}\label{conflict-of-interest}

All authors are current or former employees of LookDeep Health. This affiliation is disclosed as a potential conflict of interest.

\section{Funding}\label{funding}

This analysis was conducted as part of internal quality and research operations at LookDeep Health; no external grant funding supported this study.

\section{Author Contributions (CRediT)}\label{author-contributions-credit}

PG: Writing - original draft, Writing - review \& editing, Conceptualization, Data curation, Formal analysis, Investigation, Methodology, Project administration, Software, Validation, Visualization.

PR: Data curation, Investigation, Visualization, Writing - review \& editing.

ZD: Data annotation.

TT: Methodology, Resources, Software, Validation, Writing - review \& editing.

TW: Investigation, Resources, Writing - review \& editing.

NS: Funding acquisition, Project administration, Supervision, Writing - review \& editing.

\section{Data Availability}\label{data-availability}

De-identified analysis scripts and aggregate output tables are available upon reasonable request to the corresponding author. Individual-level monitoring data cannot be shared publicly due to the data use agreement governing this study. Reviewer-facing reproducibility notes, a data dictionary, and an artifact map accompany the submission packet.

\section{Ethical Considerations}\label{ethical-considerations}

The data used in this retrospective study was collected from patients admitted to one of eleven hospital partners across three different states in the USA. The study and handling of data followed the guidelines provided by CHAI standards. Access to this data was granted to the researchers through a Business Associate Agreement (BAA) specifically for monitoring patients at high risk of falls. In compliance with the Health Insurance Portability and Accountability Act (HIPAA), patients provided written informed consent for monitoring as part of their standard inpatient care. To ensure patient privacy, all video data was blurred prior to storage, and no identifiable information is included in this work. Face-blurred frames were used only for training purposes. Faces were manually labeled on fully-blurred images, and the raw images were then treated with a local Gaussian blur in the facial regions, ensuring privacy without compromising model training quality. The outcomes of this analysis did not influence patient care or clinical outcomes.

\section{Appendix}\label{appendix}

\textbf{Table A1. Fall mechanism taxonomy by pre-fall location (observation cohort, n=32 events).}

{\def\LTcaptype{none} 
\begin{longtable}[]{@{}
  >{\raggedright\arraybackslash}p{(\linewidth - 8\tabcolsep) * \real{0.3300}}
  >{\raggedright\arraybackslash}p{(\linewidth - 8\tabcolsep) * \real{0.1800}}
  >{\raggedright\arraybackslash}p{(\linewidth - 8\tabcolsep) * \real{0.1700}}
  >{\raggedright\arraybackslash}p{(\linewidth - 8\tabcolsep) * \real{0.1800}}
  >{\raggedright\arraybackslash}p{(\linewidth - 8\tabcolsep) * \real{0.1200}}@{}}
\toprule\noalign{}
\endhead
\bottomrule\noalign{}
\endlastfoot
\textbf{Mechanism category} & \textbf{Chair (n=7)} & \textbf{Bed (n=12)} & \textbf{Room (n=13)} & \textbf{Total} \\
Footrest / positioning failure & 6 (86\%) & --- & --- & 6 \\
Transfer failure & --- & 6 (50\%) & 2 (15\%) & 8 \\
Other / unclassified & 1 (14\%) & 6 (50\%) & 11 (85\%) & 18 \\
\end{longtable}
}

\textbf{Table A2. Event classification by furniture-origin chain (observation cohort, n=32).}

{\def\LTcaptype{none} 
\begin{longtable}[]{@{}
  >{\raggedright\arraybackslash}p{(\linewidth - 4\tabcolsep) * \real{0.2917}}
  >{\raggedright\arraybackslash}p{(\linewidth - 4\tabcolsep) * \real{0.1806}}
  >{\raggedright\arraybackslash}p{(\linewidth - 4\tabcolsep) * \real{0.2361}}@{}}
\toprule\noalign{}
\endhead
\bottomrule\noalign{}
\endlastfoot
\textbf{Chain Category} & \textbf{Events} & \textbf{\% of Total} \\
direct\_bed & 12 & 37.5\% \\
bed\_origin\_room & 10 & 31.3\% \\
direct\_chair & 7 & 21.9\% \\
chair\_origin\_room & 3 & 9.4\% \\
\end{longtable}
}

\textbf{Table A3. Multimodal LLM fall-detection performance (3-way overlap subset, n=42 visible primary rows).}

{\def\LTcaptype{none} 
\begin{longtable}[]{@{}
  >{\raggedright\arraybackslash}p{(\linewidth - 8\tabcolsep) * \real{0.1838}}
  >{\raggedright\arraybackslash}p{(\linewidth - 8\tabcolsep) * \real{0.1691}}
  >{\raggedright\arraybackslash}p{(\linewidth - 8\tabcolsep) * \real{0.1985}}
  >{\raggedright\arraybackslash}p{(\linewidth - 8\tabcolsep) * \real{0.2574}}
  >{\raggedright\arraybackslash}p{(\linewidth - 8\tabcolsep) * \real{0.1765}}@{}}
\toprule\noalign{}
\endhead
\bottomrule\noalign{}
\endlastfoot
\textbf{Model} & \textbf{Fall Sensitivity} & \textbf{Non-fall Specificity} & \textbf{Location Accuracy (detected)} & \textbf{Fall-time MAE (s)} \\
Gemini 2.5 Flash & 0.842 & 0.750 & 0.531 & 1,199 \\
Gemini 2.5 Pro & 0.684 & 1.000 & 0.654 & 521 \\
Gemini 3.1 Pro Preview & 0.447 & 1.000 & 0.588 & 106 \\
\end{longtable}
}

\section{References}\label{references}

\protect\phantomsection\label{refs}

\end{document}